\documentclass[12pt]{article}

\newif\iffinal
\finaltrue

\usepackage{sbc-template}
\usepackage{graphicx}
\usepackage[hyphens]{url}
\usepackage[T1]{fontenc} %
\usepackage[utf8]{inputenc}
\usepackage[brazil]{babel}
\usepackage[dvipsnames,table, xcdraw]{xcolor}
\usepackage{booktabs}
\usepackage{amsmath}
\usepackage{float}
\usepackage{array}
\usepackage[capitalise,noabbrev]{cleveref}
\usepackage{ifthen}
\usepackage{xcolor}
\usepackage[Algoritmo]{algorithm}
\usepackage{algpseudocode}

\usepackage{soul}
\setuldepth{Berlin}
\usepackage[authoryear,square]{natbib}
\newcommand*{\RL}[2][]{\textcolor{Rhodamine}{[\textbf{\ifthenelse{\equal{#1}{}}{RL}{RL(#1)}}: #2]}}

\title{Além do Desempenho: Um Estudo da Confiabilidade de Detectores de Deepfakes}

\iffinal
    \author{Lucas Lopes\inst{1}, Rayson Laroca\inst{2,1}, André Grégio\inst{1}
    }

    \address{Universidade Federal do Paraná
      (UFPR), Curitiba, Brasil\\
      \inst{2}\hspace{0.25mm}Pontifícia Universidade Católica do Paraná (PUCPR), Curitiba, Brasil\\[1.5ex]
      \tt\small \inst{1}\hspace{-0.2mm}\{lflopes,gregio\}@inf.ufpr.br \quad \inst{2}\hspace{0.15mm}rayson@ppgia.pucpr.br
    }
\else
    \author{Autores Anônimos}

    \address{Afiliação Anônima}
\fi

\newcommand\blue[1]{{#1}} %

\begin{document} 

\maketitle

\begin{abstract}
Deepfakes are synthetic media generated by artificial intelligence, with positive applications in education and creativity, but also serious negative impacts such as fraud, misinformation, and privacy violations. Although detection techniques have advanced, comprehensive evaluation methods that go beyond classification performance remain lacking. This paper proposes a reliability assessment framework based on four pillars: transferability, robustness, interpretability, and computational efficiency. An analysis of five state-of-the-art methods revealed significant progress as well as critical limitations. 
\end{abstract}
     
\begin{resumo} 
Deepfakes são mídias sintéticas geradas por inteligência artificial, com aplicações positivas na educação e na criatividade, mas também com impactos negativos graves, como fraudes, desinformação e violações de privacidade.
Apesar dos avanços em técnicas de detecção, ainda há escassez de métodos de avaliação abrangentes que considerem aspectos além do desempenho em classificação. Este trabalho propõe um framework de avaliação da confiabilidade baseado em quatro pilares: transferibilidade, robustez, interpretabilidade e eficiência computacional.
A análise de cinco métodos do estado da arte revelou avanços significativos, mas também limitações críticas.
\end{resumo}

\section{Introdução}

\begingroup
    \renewcommand\thefootnote{}
    Nos últimos anos, a geração de conteúdo sintético por inteligência artificial~(IA) tem ganhado destaque tanto na academia quanto na indústria~\citep{abdullah:24}.
    Dentre as diversas tecnologias emergentes, os \textit{deepfakes} se destacam pela capacidade de produzir imagens e vídeos altamente realistas~\citep{pei:24}.
    Embora essa tecnologia possua aplicações legítimas nos ramos de entretenimento, educação e comércio~\citep{caporusso:21,roe:24}, sua disseminação também levanta sérias preocupações no âmbito social, político e financeiro, especialmente quando utilizada para fraudes, desinformação e ataques à privacidade~\citep{wang:24:deepfake}.\footnote{Artigo aceito no Simpósio Brasileiro de Cibersegurança (SBSeg 2025). A versão publicada está disponível na SBC-OpenLib~(SOL) (\url{https://doi.org/10.5753/sbseg.2025.11431}).}
    \addtocounter{footnote}{-1}
\endgroup

O termo deepfake se refere a uma técnica de síntese de mídia baseada em redes neurais profundas.
Desde sua popularização em 2017, com um modelo de troca de rostos (\textit{face swap}) divulgado no Reddit~\citep{deepfakes:17}, a tecnologia evoluiu significativamente, passando a permitir reencenação facial, modificação de expressões, e até geração completa de vídeos reencenados~\citep{suwajanakorn:17}. No cinema, deepfakes foram utilizados para rejuvenescer atores em filmes como \textit{Gemini Man}~\citep{murphy:23}, enquanto na educação tutores virtuais personalizados foram criados com IA para atender diferentes perfis de alunos~\citep{roe:24}.
No setor de varejo, marcas já exploram o uso da tecnologia para experimentação virtual de roupas e acessórios~\citep{caporusso:21}. 

Contudo, o uso indevido de deepfakes tem crescido de maneira alarmante. Em fevereiro de 2024, um golpe utilizando deepfakes em uma videoconferência levou uma multinacional de Hong Kong a perder 25 milhões de dólares~\citep{melappalayam:24}. Em abril do mesmo ano, um fraudador se passando por Elon Musk enganou uma vítima, resultando no desvio do equivalente a 50 mil dólares~\citep{sharma:24}. Esses eventos demonstram o avanço das técnicas de manipulação e a urgência do desenvolvimento de sistemas de detecção robustos e confiáveis. 

\blue{Diante desse cenário, o desenvolvimento de técnicas eficazes para a detecção de deepfakes tornou-se essencial. Diversos métodos têm sido propostos, com abordagens que vão desde o uso de características físicas faciais~\citep{zhou:21, ortega:22, wang:24:deepfake} até arquiteturas sofisticadas baseadas em aprendizado profundo~\citep{jeong:22, kong:24}. No entanto, observa-se que grande parte desses trabalhos concentra sua avaliação em métricas tradicionais de desempenho, como acurácia~(ACC) ou o valor da área sob a curva ROC~(AUC) em conjuntos de teste padronizados, sem considerar aspectos críticos como robustez a perturbações, interpretabilidade dos resultados, ou viabilidade~computacional.}

\blue{Neste trabalho, propomos uma abordagem mais abrangente para a avaliação de detectores de deepfakes, estruturada em quatro pilares fundamentais: transferibilidade, robustez, interpretabilidade e eficiência computacional. Além disso, oferecemos uma quantificação objetiva e padronizada desses pilares, consolidando-os em um escore de confiabilidade global que permite comparar métodos sob uma perspectiva mais realista.}

\blue{Além disso, realizamos uma revisão dos principais métodos de geração e detecção da literatura recente, os quais foram reagrupados em categorias com base em suas estratégias e técnicas predominantes. Como prova de conceito, aplicamos o framework proposto a cinco detectores representativos do estado da arte, analisando seus pontos fortes e limitações em múltiplas dimensões de confiabilidade.} \blue{Com isso, esperamos contribuir para uma avaliação mais ampla e crítica dos métodos de detecção de deepfakes, estimulando o desenvolvimento de soluções mais transparentes, robustas e~acessíveis.}

\section{Fundamentos e Tecnologias de Geração de Deepfakes}
\label{sec:fundamentos}

As técnicas de geração de deepfakes podem ser agrupadas em três categorias principais: troca de rostos~(\textit{face swap}), que substitui a identidade facial em imagens ou vídeos~\citep{deepfakes:17}; recriação de expressões~(\textit{face reenactment}), que transfere gestos e emoções de um indivíduo para outro~\citep{suwajanakorn:17}; e geração de rostos falantes~(\textit{talking face generation}), que sincroniza imagens faciais com áudio~\citep{xie:24}.

O avanço dessas técnicas está fortemente associado à evolução de modelos generativos~\citep{pei:24}, especialmente \textit{autoencoders}, redes adversárias generativas~(GANs, do Inglês \textit{Generative Adversarial Networks})~\citep{goodfellow:14} e, mais adiante, modelos de difusão (DDPM, do Inglês \textit{Denoising Diffusion Probabilistic Models})~\citep{ho:20}, que possibilitam a síntese de imagens realistas com diferentes graus de controle, fidelidade e qualidade~\citep{santos2024multi}.

\subsection{Autoencoders}

\begin{figure}[!htb]
    \centering
    \includegraphics[width=0.9\linewidth]{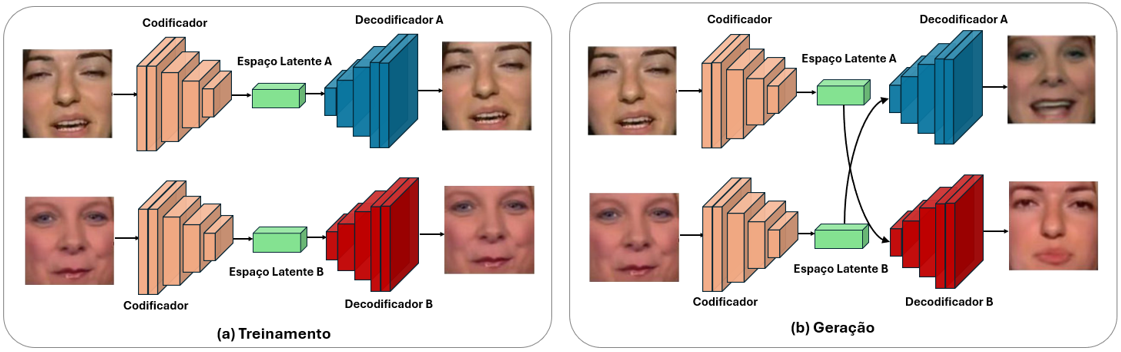}
    \vspace{-1mm}
    \caption{Esquema de uso de autoencoders para reencenação: (a)~Treinamento com codificador compartilhado e decodificadores específicos por identidade; (b)~Geração da face final, combinando a expressão da imagem de entrada com a identidade do alvo.
}
    \label{fig:vae}
\end{figure}

Autoencoders foram amplamente utilizados nos primeiros modelos de \textit{face swap} e reencenação, com um codificador compartilhado e decodificadores específicos para cada identidade. Como ilustrado na Figura~\ref{fig:vae}, essa arquitetura permite reconstruir a expressão facial original com a identidade do rosto-alvo~\citep{deepfakes:17,wang:24:deepfake}.
No entanto, exigem decodificadores treinados por identidade, o que limita sua escalabilidade. Para contornar essa limitação, modelos independentes de identidade passaram a incorporar GANs no processo de geração.

\subsection{Redes Adversárias Generativas (GANs)}

\begin{figure}[!htb]
    \centering
    \includegraphics[width=0.775\textwidth]{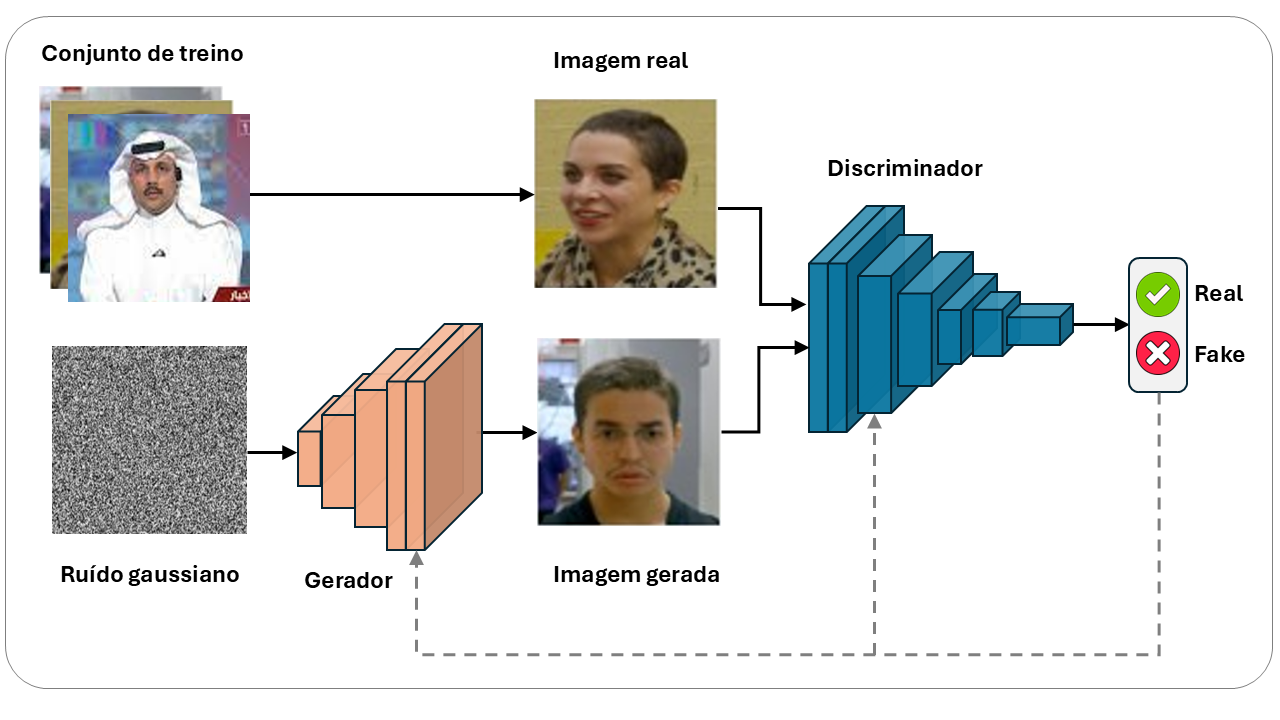}

    \vspace{-3mm}
    
    \caption{Esquema de treinamento de uma GAN para geração de faces. O gerador cria imagens sintéticas enquanto o discriminador tenta distingui-las das reais, em um processo adversarial que leva à geração de faces mais~realistas.}
    \label{fig:gan}
\end{figure}

As GANs, compostas por um gerador e um discriminador treinados de forma adversarial~(veja a Figura~\ref{fig:gan}),
tornaram-se o principal mecanismo para a geração de faces realistas~\citep{goodfellow:14}.
Abordagens como FaceShifter~\citep{li:19} e SimSwap~\citep{chen:21} propõem modelos independentes de identidade que integram características faciais da origem na imagem-alvo, com maior robustez a oclusões.
Posteriormente, arquiteturas baseadas no StyleGAN~\citep{karras:21} viabilizaram a geração de imagens em alta resolução~\citep{wang:24:deepfake}.
Apesar da alta qualidade dos resultados, o treinamento de GANs exige cuidados com a estabilidade e a escolha de hiperparâmetros, além de etapas adicionais de pós-processamento para a remoção de artefatos~\citep{kim:25}.

\subsection{Modelos de Difusão}

\begin{figure}[!htb]
    \centering
    \includegraphics[width=0.775\textwidth]{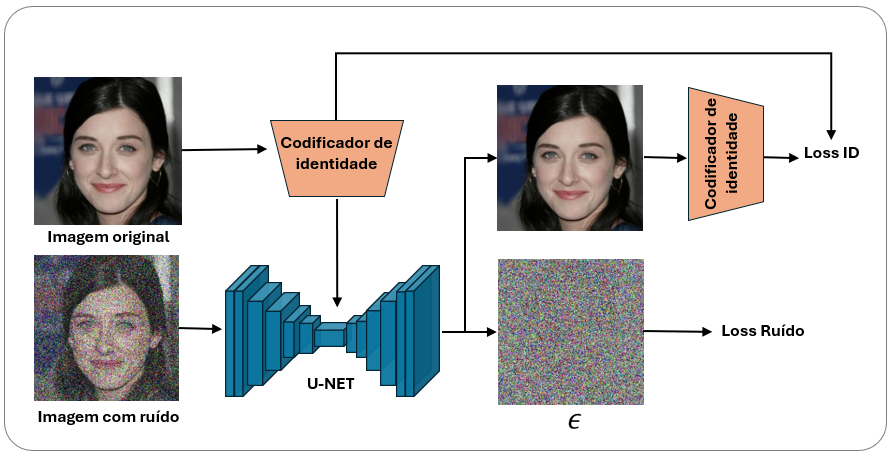}
    \vspace{-1mm}
    \caption{Esquema de treinamento do DiffFace. Um codificador de identidade insere atributos no espaço latente da U-Net, que aprende a reconstruir rostos a partir de imagens ruidosas. As perdas de ruído e identidade orientam o ajuste dos pesos. Adaptado de~\cite{kim:25}.}
    \label{fig:dm}
\end{figure}

Modelos de difusão surgem como alternativa às GANs, oferecendo treinamento mais estável e maior diversidade amostral, ao gerar imagens por meio da reversão de um processo de adição de ruído~\citep{bishop:24}.
Métodos como DiffFace~\citep{kim:25} e DiffSwap~\citep{zhao:23} aplicam esse princípio à tarefa de geração e troca de rostos, tratando-a como um problema de \textit{inpainting} condicional, com foco na preservação da identidade e expressividade do alvo. A Figura~\ref{fig:dm} ilustra o processo de treinamento de um modelo de difusão para geração de faces.

Outro exemplo notável é o \textit{X-Portrait}~\citep{xie:24}, que permite a geração de animações altamente realistas a partir de imagens estáticas.
Apesar do potencial, esses modelos ainda enfrentam limitações relacionadas ao alto custo computacional e ao tempo de inferência, decorrentes das múltiplas etapas de denoising exigidas para gerar amostras de alta fidelidade~\citep{bishop:24,shen:25:efficient}. 

\section{Panorama dos Métodos de Detecção}
\label{sec:panorama}

Diversas abordagens têm sido propostas para a detecção de deepfakes, geralmente classificadas em três categorias~\citep{moura:21}: (i)~métodos baseados em inconsistências físicas, (ii)~abordagens orientadas por dados, e (iii)~técnicas focadas em artefatos de síntese. Neste trabalho, introduzimos uma quarta categoria --~ (iv) \textit{métodos híbridos}, que combinam múltiplas estratégias.

\subsection{Inconsistências Físicas}

Deepfakes frequentemente introduzem discrepâncias fisiológicas sutis que, embora imperceptíveis a olho nu, podem ser exploradas para sua detecção.
Entre essas anomalias, destacam-se padrões irregulares no piscar dos olhos, inconsistentes na taxa de batimentos cardíacos inferidas pela coloração da pele, e movimentos labiais dessincronizados em relação ao áudio.
Abordagens como a proposta por~\cite{li:18} utilizaram CNNs combinadas com redes neurais recorrentes~(RNNs, do inglês \textit{Recurrent Neural Networks}), formando uma arquitetura denominada LRCN~(\textit{Long-term Recurrent Convolutional Network}), para capturar tanto as correlações espaciais --~associadas ao estado de abertura dos olhos~-- quanto as dinâmicas temporais ao longo de sequências de vídeo.

Avançando na análise espaço-temporal,~\cite{ortega:22} propuseram a detecção de deepfakes baseada em variações sutis de coloração facial, associadas ao fluxo sanguíneo.
A abordagem utiliza uma técnica de fotopletismografia remota para identificar padrões de pulsação natural que estão ausentes ou apresentam inconsistências temporais em mídias sintéticas.

Ainda nesta categoria, métodos multimodais têm explorado a dessincronização entre as modalidades visual e auditiva como um indício de manipulação.
\cite{zhou:21} propuseram a integração de \textit{embeddings} extraídos de quadros faciais e espectrogramas de áudio por meio de mecanismos de atenção cruzada, possibilitando a detecção de inconsistências entre os movimentos labiais e o conteúdo sonoro.
Esses avanços evidenciam que a análise de sinais fisiológicos e sua coerência temporal desempenha um papel fundamental na melhoria da eficácia dos sistemas de~detecção.

\subsection{Modelos Orientados por Dados}

Nesta categoria, redes neurais profundas --~comumente CNNs e redes de memória de curto e longo prazo (LSTM, do inglês \textit{Long Short-Term Memory})~-- são treinadas diretamente em grandes volumes de dados reais e sintéticos.
Um exemplo recente é o trabalho de~\cite{ezeakunne:24}, que propuseram a geração de amostras auto-misturadas para compor conjuntos de dados com maior diversidade demográfica, favorecendo a robustez do modelo frente a diferentes perfis populacionais.
A arquitetura adotada utiliza a EfficientNet \citep{tan:19} como extratora de características e dois classificadores do tipo perceptron multicamada~(MLP, do inglês \textit{Multi-Layer Perceptron}): um para a detecção binária de deepfakes e outro para a predição do grupo demográfico. O treinamento é conduzido de forma multi-tarefa, com otimização conjunta das perdas associadas a ambas as saídas, promovendo o aprendizado de representações discriminativas menos sensíveis a variações~demográficas.

Um segundo exemplo é a CLRNet~\citep{tariq:21}, que utiliza uma arquitetura baseada em uma rede convolucional residual acoplada a células LSTM, capazes de capturar simultaneamente padrões espaciais e temporais em sequências de vídeo. Explorando diferentes estratégias de treinamento, como aprendizado em domínio único, fusão de domínios e transferência de aprendizado, a CLRNet demonstrou alta capacidade de generalização, mesmo frente a conjuntos de dados desafiadores como o DeepFake-in-the-Wild~\citep{zi:20:wilddeepfake}.

Mais recentemente, esta categoria passou a contar com abordagens de detecção baseadas em aprendizado não supervisionado. Nesta perspectiva, \cite{diniz:24} propuseram uma abordagem de detecção open-set, onde apenas amostras genuínas são usadas no treinamento. Para facilitar a separação entre padrões reais e falsificados no espaço latente, foi aplicada a função de perda triplet loss \citep{schroff:15} com mineração de exemplos difíceis. Utilizando classificadores leves, como One-Class SVM, Isolation Forest e Extreme Value Machine, o método alcançou AUC superior a 0,80, sem exposição prévia a exemplos falsificados.

\subsection{Detecção por Artefatos}

A geração sintética comumente introduz artefatos específicos, como distorções, desfoques ou padrões anômalos em domínios frequenciais --~sinais que podem ser explorados por métodos de detecção.
O método FWA (\textit{Face Warping Artifacts}), proposto por~\citep{li:19:exposing}, por exemplo, explora deformações faciais decorrentes da etapa final de redimensionamento aplicada às faces sintetizadas.
Essa operação, utilizada para alinhar a face gerada ao rosto-alvo, gera inconsistências de resolução entre a região facial e as áreas adjacentes.
Para detectar tais artefatos, o FWA emprega CNNs treinadas com dados sintéticos obtidos por meio de simples transformações geométricas, dispensando o uso de exemplos negativos (deepfakes) durante o~treinamento.

Outra abordagem de detecção por artefatos atua no domínio da frequência, utilizando transformadas espectrais para capturar padrões sutis introduzidos durante o processo de síntese.
\cite{giudice:21} propuseram a técnica CTF-DCT~(\textit{Capture The Fake via Discrete Cosine Transform}), que identifica assinaturas específicas de redes geradoras --~denominadas Frequências Específicas de GAN~-- por meio da análise da distribuição dos coeficientes AC da Transformada Discreta do Cosseno~(DCT, do inglês \textit{Discrete Cosine Transform}) aplicada a blocos de 8$\times$8~pixels.
A abordagem estima estatísticas baseadas em distribuições Laplacianas para detectar anomalias de forma robusta, mesmo na presença de perturbações como compressão JPEG, espelhamento ou~rotação.

Com o avanço dos modelos de difusão \citep{shen:25:efficient}, tornaram-se necessárias novas abordagens para detectar artefatos mais refinados.
O método Synthbuster~\citep{bammey:24} introduz uma metodologia baseada em análise espectral de resíduos de alta frequência.
Inicialmente, é aplicado um filtro de diferença cruzada para ressaltar componentes de alta frequência, seguido da análise da Transformada Rápida de Fourier~(FFT, do inglês \textit{Fast Fourier Transform}) para extrair magnitudes específicas associadas a períodos de 2, 4 e 8 pixels --~padrões característicos da etapa de amostragem e \textit{upscaling} dos modelos de difusão.
Assim, um classificador leve baseado em \textit{Gradient Boosting Trees} é então treinado apenas utilizando os picos de magnitude extraídos, alcançando alta acurácia mesmo em imagens comprimidas e demonstrando capacidade de generalização para modelos de geração não vistos durante o~treinamento.

\subsection{Métodos Híbridos}

Métodos híbridos de detecção de deepfakes têm sido desenvolvidos com o objetivo de combinar os pontos fortes de diferentes abordagens, resultando em modelos mais resilientes.
O método FrePGAN~\citep{jeong:22}, por exemplo, utiliza GANs para aplicar perturbações no domínio da frequência em ambas as classes (real e fake) durante o treinamento, reduzindo o \textit{overfitting} a artefatos específicos de determinados geradores.
De forma complementar, o FreqNET~\citep{tan:24} propõe um módulo integrado de aprendizado no domínio espacial e espectral, baseado em uma camada convolucional de Fourier, projetada para capturar tanto a distribuição de magnitude quanto de fase das componentes da Transformada de Fourier, demonstrando alta eficácia na detecção de mídias sintéticas geradas por 17 tipos distintos de~GANs.

Outro exemplo relevante é o método OSDFD~\citep{kong:24} que emprega uma arquitetura baseada em Transformers visuais~(ViT), combinada com LoRA --~uma técnica leve de ajuste fino que reduz os custos de adaptação de grandes modelos.
O método incorpora ainda a estratégia de mistura de estilos de falsificação~(\textit{forgery style mixture}), promovendo maior diversidade no espaço latente durante o treinamento e aprimorando a capacidade do detector de lidar com técnicas de falsificação inéditas.
Esses avanços ilustram o potencial das abordagens combinatórias na construção de detectores de deepfakes mais robustos e generalizáveis.

\section{Métodos de Prevenção a Deepfakes}
\label{sec:prevencao}

\blue{Além das estratégias de detecção discutidas neste trabalho, também foram propostas abordagens preventivas contra o uso malicioso de deepfakes. Uma das mais exploradas é a \textit{deepfake destruction}, que insere perturbações cuidadosamente calculadas nas imagens para desestabilizar o processo de geração, impedindo a criação eficaz de falsificações \citep{ruiz:20}. Outra linha promissora é a \textit{scapegoat generation} \citep{kato:23}, que modifica adversarialmente o rosto original com o objetivo de produzir uma imagem reconhecível pelo usuário como um avatar, mas com identidade suficientemente alterada para inviabilizar sua utilização por modelos de~síntese.}

\blue{Mais recentemente, destacam-se iniciativas de marcação digital invisível, como o SynthID, desenvolvido pelo Google DeepMind para embutir marcas d'água imperceptíveis em conteúdos gerados por IA. Em parceria com a NVIDIA, o SynthID passou a ser integrado aos modelos generativos na plataforma \textit{build.nvidia.com}, ampliando sua aplicação a diferentes ecossistemas de IA \citep{deepmind:25}. Essas estratégias reforçam a importância de uma abordagem integrada que combine prevenção e detecção, a fim de mitigar os impactos sociais e informacionais dos deepfakes em escala.}

\section{Confiabilidade na Detecção de Deepfakes}
\label{sec:confiabilidade}

Apesar dos avanços recentes nos mecanismos de defesa, os métodos de detecção ainda estão longe de alcançar um desempenho confiável em cenários não controlados~\citep{wang:24:deepfake}.
Muitos modelos apresentam resultados expressivos em ambientes laboratoriais, mas demonstram queda de desempenho ao lidar com dados provenientes de diferentes origens, resoluções ou técnicas de manipulação~\citep{pei:24}. 

Mais do que considerar métricas isoladas, como acurácia ou AUC, é fundamental avaliar a confiabilidade dos detectores, entendida como a capacidade de manter um desempenho estável e robusto em condições variadas.
Conforme proposto por~\cite{wang:24:deepfake}, três pilares centrais orientam essa avaliação: transferibilidade, robustez e interpretabilidade.
Neste trabalho, propomos inclusão de um quarto pilar, a \textit{eficiência computacional}, visando contemplar também a viabilidade prática dos modelos em ambientes com restrições de tempo e~recursos.

Esses quatro pilares fornecem uma base abrangente para uma avaliação alinhada com cenários realistas, permitindo que os modelos de detecção sejam analisados não apenas em termos de precisão, mas também sob outras dimensões cruciais de~confiabilidade.

\subsection{Transferibilidade}

Transferibilidade refere-se à capacidade dos modelos de generalizarem para dados não vistos, especialmente aqueles gerados por técnicas diferentes daquelas utilizadas durante o treinamento~\citep{abdullah:24, wang:24:deepfake}. Detectores que dependem fortemente de artefatos específicos dos conjuntos de dados tendem a sofrer quedas significativas de desempenho em cenários de avaliação entre datasets distintos~(\textit{cross-dataset})~\citep{abdullah:24}.
Para mitigar esse problema, têm sido exploradas estratégias como o aumento de dados com mapas de perturbação~\citep{jeong:22} e a mistura de estilos entre diferentes tipos de deepfakes~\citep{kong:24}.
Embora esses métodos apresentem resultados promissores, eles ainda se mostram insuficientes para assegurar uma transferibilidade ampla e consistente.

\subsection{Robustez}

Robustez diz respeito à capacidade do modelo de manter seu desempenho diante de perturbações naturais ou artificiais, como compressão, ruídos, distorções ou ataques adversariais~\citep{wang:24:deepfake}.
Estudos mostram que detectores treinados em condições ideais tendem a sofrer quedas significativas de desempenho quando expostos a cenários de baixa qualidade ou manipulações maliciosas.
\cite{abdullah:24}, por exemplo, avaliaram oito métodos de detecção sob ataques adversariais baseados em gradiente aplicados diretamente à entrada, evidenciando a fragilidade de todos os modelos frente a esse tipo de perturbação.
Em resposta, exploram-se estratégias como o treinamento com dados degradados \citep{jeong:22}, o uso de representações mais robustas {\citep{kong:24}, e o desenvolvimento de arquiteturas mais resilientes à compressão~\citep{kundu:25}.

\subsection{Interpretabilidade}

A interpretabilidade é essencial em contextos críticos, como segurança, processos jurídicos e perícia digital \citep{wang:24:deepfake}.
Nesse contexto, ela diz respeito à capacidade de explicar e justificar as decisões tomadas por um algoritmo --~por exemplo, apontar que uma imagem foi classificada como falsa devido à presença de artefatos anômalos em determinadas regiões, como bordas borradas ou padrões de iluminação inconsistentes \citep{trinh:21}.
Modelos de detecção baseados em redes neurais profundas, por exemplo, são frequentemente criticados por sua natureza de “caixa-preta”, pois não oferecem uma compreensão clara dos fatores que motivaram determinada decisão~\citep{pedulla:22}.

Nesta direção, algumas abordagens têm explorado o uso de mapas de saliência~\citep{pontorno:24} e análises espectrais~\citep{bammey:24} para explicar o comportamento dos modelos.
No entanto, essas soluções ainda enfrentam desafios relacionados à clareza, consistência e padronização das explicações fornecidas.
Um dos métodos mais promissores em termos de interpretabilidade foi proposto por~\cite{kundu:25}, que utilizaram uma arquitetura multimodal baseada em modelos de linguagem natural para gerar explicações em linguagem humana sobre as decisões de~classificação.

\subsection{Eficiência Computacional}

Eficiência computacional envolve a capacidade de realizar inferências rápidas com uso moderado de recursos, aspecto fundamental para aplicações em tempo real ou em dispositivos com restrições de hardware~\citep{wang:24:computation}. No entanto, detectores de alta capacidade, baseados em transformers~\citep{kong:24, kundu:25} ou ensembles~\citep{zhou:21}, frequentemente demandam alto poder computacional. Dessa forma, arquiteturas neurais mais leves, como no método proposto por~\cite{luo:24}, que utiliza a EfficientNet-B4 como classificador, buscam equilibrar capacidade de classificação e eficiência~computacional.

\section{Proposta de Framework de Avaliação}
\label{sec:framework}

A aplicação prática de detectores de deepfakes exige mais do que alta acurácia em ambientes controlados --~requer também confiabilidade sob diferentes condições e restrições operacionais~\citep{wang:24:deepfake}.
Embora a maioria dos métodos citados neste trabalho concentrem suas avaliações em métricas tradicionais, como o desempenho em testes \textit{cross-dataset} e eventuais avaliações de robustez, aspectos críticos como interpretabilidade e eficiência computacional permanecem, em grande parte, negligenciados.

\blue{Com base nessa lacuna, propomos um framework de avaliação estruturado nos quatro pilares fundamentais: transferibilidade, robustez, interpretabilidade e eficiência computacional, conforme detalhado na seção anterior. 
Diferentemente de abordagens existentes, nossa proposta oferece uma \textit{quantificação formal e integrada} desses quatro pilares, permitindo comparar modelos sob uma perspectiva multidimensional de confiabilidade.}
A Figura~\ref{fig:confialibidade} ilustra a distinção entre a abordagem tradicional de avaliação e a metodologia abrangente aqui~proposta.

\begin{figure}[!htb]
    \centering
    \includegraphics[width=0.875\linewidth]{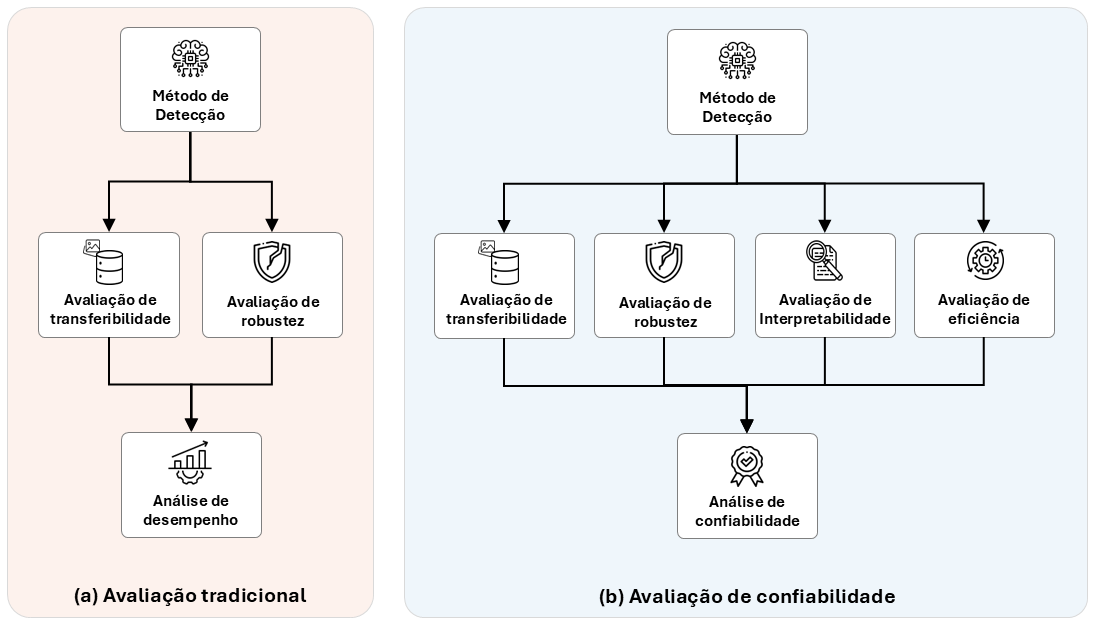}
    \vspace{-1mm}
    \caption{Comparação entre (a)~a abordagem tradicional de avaliação de detectores de deepfakes, centrada apenas em métricas de desempenho e robustez, e (b)~o framework proposto neste trabalho, que incorpora e quantifica os quatro pilares da confiabilidade.}
    \label{fig:confialibidade}
\end{figure}

\blue{Cada pilar é mensurado por métricas específicas e complementares. Para os dois primeiros (transferibilidade e robustez) optamos por permitir o uso de AUC ou ACC, dada a ausência de consenso na literatura. Essa escolha visa compatibilidade entre diferentes práticas experimentais, mantendo a comparabilidade entre métodos.}

A \textbf{transferibilidade} é avaliada pela média da AUC ou ACC obtida em testes \textit{cross-dataset}, refletindo a capacidade do modelo de generalizar para dados oriundos de diferentes técnicas e domínios. Formalmente, seja \( N \) o número de conjuntos de dados não vistos usados nos testes, a métrica de transferibilidade \( T \) é definida pela Equação~\ref{eq:transferibilidade}:
\begin{equation}
T = \frac{1}{N} \sum_{i=1}^{N} \text{Score}_{\text{cross}}(i) \, .
\label{eq:transferibilidade}
\end{equation}

De forma análoga, a \textbf{robustez} considera o desempenho médio do modelo frente a três categorias de perturbações: compressão, ruído e ataques adversariais. Para garantir uma avaliação equilibrada entre esses tipos de degradação, a métrica de robustez \( R \) é definida como a média das médias obtidas em cada grupo de testes, conforme a Equação~\ref{eq:robustez}:
\begin{equation}
R = \frac{1}{3} \left( \frac{1}{C} \sum_{i=1}^{C} \text{Score}_{\text{comp}}(i) 
+ \frac{1}{P} \sum_{j=1}^{P} \text{Score}_{\text{perturb}}(j) 
+ \frac{1}{A} \sum_{k=1}^{A} \text{Score}_{\text{adv}}(k) \right) \, ,
\label{eq:robustez}
\end{equation}
\noindent onde:
\begin{itemize}
    \item \( C \), \( P \) e \( A \) representam, respectivamente, o número de testes realizados sob compressão, perturbações com ruídos, e ataques adversariais;
    \item \( \text{Score}_{\text{comp}}(i) \) é a AUC ou ACC obtida no \( i \)-ésimo teste com compressão;
    \item \( \text{Score}_{\text{perturb}}(j) \) é a AUC ou ACC obtida no \( j \)-ésimo teste com perturbações por~ruídos;
    \item \( \text{Score}_{\text{adv}}(k) \) é a AUC ou ACC obtida no \( k \)-ésimo teste com ataque adversarial.
\end{itemize}

A \textbf{interpretabilidade} é quantificada segundo uma escala qualitativa, apresentada na Tabela~\ref{tab:interpretabilidade}, que atribui valores de 0 a 1 com base no grau de profundidade e na integração dos mecanismos explicativos ao~modelo.

\begin{table}[!htb]
\centering
\setlength{\tabcolsep}{10pt}
\caption{Critérios de quantificação para interpretabilidade.}
\label{tab:interpretabilidade}
\renewcommand{\arraystretch}{1.2}

\vspace{-2mm}
\resizebox{0.8\linewidth}{!}{
\begin{tabular}{|l|c|m{6cm}|}
\hline
\textbf{Critério} & \textbf{Valor} & \textbf{Descrição} \\ \hline
Nenhuma explicação & 0.0 & Modelo “caixa-preta”, sem qualquer visualização ou justificativa. \\ \hline
Visualizações básicas & 0.3 – 0.5 & Uso de técnicas como Grad-CAM (mapas de saliência)  ou t-SNE, sem análise crítica. \\ \hline
Análises interpretativas & 0.6 – 0.8 & Aplicação de métodos como LIME ou SHAP, com explicações baseadas em atributos. \\ \hline
Explicabilidade integrada & 0.9 – 1.0 & \blue{Mecanismos explicativos integrados ao modelo} \\ %
\hline
\end{tabular}
}
\end{table}

\blue{A \textbf{eficiência computacional} é avaliada com base no número total de parâmetros do modelo ($P$), por ser um indicador atrelado ao custo de inferência e uso de recursos computacionais~\citep{wang:24:computation}. Essa escolha se justifica por dois fatores: (i) a ausência de código disponibilizado publicamente para os métodos analisados, o que inviabiliza medições consistentes de tempo de inferência; e (ii) a correlação entre $P$ e variáveis como consumo de memória, latência e viabilidade de execução em tempo real.}

A atribuição de escores segue a escala discretizada apresentada a seguir:
\begin{equation}
\label{eq:eficiencia}
E =
\begin{cases}
1.0, & \text{se } P < 10^7 \\
0.8, & \text{se } 10^7 \leq P < 5 \times 10^7 \\
0.6, & \text{se } 5 \times 10^7 \leq P < 10^8 \\
0.4, & \text{se } 10^8 \leq P < 3 \times 10^8 \\
0.2, & \text{se } 3 \times 10^8 \leq P < 10^9 \\
0.0, & \text{se } P \geq 10^9
\end{cases}
\end{equation}

\blue{O Algoritmo~\ref{alg:confiabilidade} formaliza o cálculo do escore de confiabilidade global ($SCG$) a partir dos quatro pilares definidos no framework. Cada dimensão é incorporada de maneira padronizada, permitindo a integração com sistemas automatizados de avaliação.
Com exceção do pilar de interpretabilidade, que ainda depende de uma classificação manual (ver Tabela~\ref{tab:interpretabilidade}), os demais pilares podem ser computados de forma objetiva a partir dos resultados experimentais.}

\begin{algorithm}[ht]
\footnotesize %
\caption{Cálculo do Escore de Confiabilidade Global ($SCG$)}
\label{alg:confiabilidade}
\begin{algorithmic}[1]
\Require
\Statex $D$: conjunto de datasets não vistos
\Statex $P_{\text{comp}}$, $P_{\text{perturb}}$, $P_{\text{adv}}$: AUC ou ACC obtidas nos testes de robustez
\Statex $P$: número total de parâmetros do modelo
\Statex $I$: escore qualitativo de interpretabilidade (ver Tabela~\ref{tab:interpretabilidade})
\Ensure Escore de confiabilidade global $SCG \in [0,1]$
\vspace{0.2cm}
\Function{CalcularTransferibilidade}{$D$}
    \State \Return $T = \frac{1}{|D|} \sum\limits_{i=1}^{|D|} \text{Score}_{\text{cross}}(i)$ \Comment{Equação~\ref{eq:transferibilidade}}
\EndFunction
\vspace{0.2cm}
\Function{CalcularRobustez}{$P_{\text{comp}}, P_{\text{perturb}}, P_{\text{adv}}$}
    \State \Return $R = \frac{1}{3} \left( \frac{1}{C} \sum\limits_{i=1}^{C} \text{Score}_{\text{comp}}(i) 
    + \frac{1}{P} \sum\limits_{j=1}^{P} \text{Score}_{\text{perturb}}(j) 
    + \frac{1}{A} \sum\limits_{k=1}^{A} \text{Score}_{\text{adv}}(k) \right)$ \Comment{Equação~\ref{eq:robustez}}
\EndFunction
\vspace{0.2cm}
\Function{CalcularEficiência}{$P$}
    \State \Return $E$ conforme a escala definida na Equação~\ref{eq:eficiencia}
\EndFunction
\vspace{0.2cm}
\Function{CalcularConfiabilidade}{$D, P_{\text{comp}}, P_{\text{perturb}}, P_{\text{adv}}, I, P$}
    \State $T \gets$ \Call{CalcularTransferibilidade}{$D$}
    \State $R \gets$ \Call{CalcularRobustez}{$P_{\text{comp}}, P_{\text{perturb}}, P_{\text{adv}}$}
    \State $E \gets$ \Call{CalcularEficiência}{$P$}
    \State $SCG \gets \frac{1}{4} (T + R + I + E)$
    \State \Return $SCG$
\EndFunction
\end{algorithmic}
\end{algorithm}

\section{Estudo de Casos: Avaliação de Métodos do Estado da Arte}
\label{sec:avaliacao}

Com base no framework proposto, analisamos cinco métodos recentes de detecção de deepfakes à luz dos pilares de confiabilidade.
A seleção considerou: (i)~o desempenho reportado em cenários \textit{cross-dataset}, como indicador da capacidade de generalização; e (ii)~a avaliação sob condições degradadas, como compressão e ruído, evidenciando testes mínimos de robustez.

\blue{Para cada método, extraímos os valores de desempenho reportados em cenários \textit{cross-dataset} e sob perturbações (compressão, ruído e ataques adversariais), priorizando AUC ou ACC, conforme a disponibilidade. Quando múltiplos testes estavam presentes, utilizamos a média aritmética, assegurando a comparabilidade. A partir disso, calculamos os escores de transferibilidade~($T$) e robustez~($R$) conforme as Equações~\ref{eq:transferibilidade} e~\ref{eq:robustez}.}

\blue{O valor de interpretabilidade~($I$) foi atribuído manualmente com base na presença, natureza e integração das explicações fornecidas, seguindo os critérios da Tabela~\ref{tab:interpretabilidade}.}

\blue{A eficiência computacional~($E$) foi estimada com base no número de parâmetros~($P$). Quando não informado, $P$ foi inferido a partir do modelo base utilizado na arquitetura (por exemplo, DINOv2 e PaliGemma2).}

\blue{Com todos os pilares quantificados, o escore de confiabilidade global~($SCG$) foi calculado conforme o Algoritmo~\ref{alg:confiabilidade}, permitindo uma avaliação mais padronizada diante da heterogeneidade dos protocolos originais.}

O \textit{\underline{FrePGAN}}~\citep{jeong:22} é um método voltado à detecção de deepfakes gerados por GANs, com foco na mitigação de \textit{overfitting} por meio da aplicação de mapas de perturbação no domínio da frequência.
Apresenta boa transferibilidade, com desempenho estável em múltiplos domínios, alcançando uma acurácia média de 0,76 em conjuntos de dados não vistos.
Também exibe robustez frente a compressão e ruído, mas não foi avaliado contra ataques adversariais, limitando sua análise completa.
A interpretabilidade é considerada baixa, uma vez que não oferece explicações claras além dos mapas espectrais. Quanto à eficiência computacional, o uso da arquitetura ResNet-50 combinada a um gerador adicional sugere um custo~moderado.

O \textit{\underline{SCLoRA}}~\citep{kong:23} utiliza uma arquitetura ViT combinada com LoRA e a função \textit{Single-Center Loss} para promover a separação entre classes.
Apresentou transferibilidade intermediária, com AUC média de 0,72, superando modelos baseados exclusivamente em ViTs.
Em termos de robustez, demonstrou desempenho consistente sob diferentes níveis de compressão (fatores c23 e c40), que simulam perdas de qualidade típicas de vídeos em redes sociais --~sendo c23 associado a compressão moderada e c40 a compressão severa, segundo o padrão do conjunto de dados FaceForensics++~\citep{rössler:19}.
Por outro lado, o modelo não foi submetido a avaliações contra ataques adversariais, o que limita a análise de sua resiliência em cenários mais hostis.
Assim como o método anterior, carece de interpretabilidade estruturada, operando como um modelo caixa-preta. Apesar do uso eficiente do LoRA, o backbone com 86M de parâmetros resulta em um custo computacional elevado durante a~inferência.

O \textit{\underline{OSDFD}}~\citep{kong:24} combina uma arquitetura ViT com módulos eficientes~(Adapter + LoRA) e técnicas de mistura de estilos de falsificação para promover maior capacidade de generalização.
Alcançou AUC média de 0,82 em seis conjuntos de dados não vistos, destacando-se em termos de transferibilidade.
Quanto à robustez, apresentou bons resultados sob diferentes níveis de compressão e perturbações, embora não tenha sido avaliado frente a ataques adversariais. Sua explicabilidade é limitada, baseada apenas no Grad-CAM~\citep{selvaraju2017gradcam}, que destaca as regiões da imagem mais relevantes para a decisão do modelo, e no t-SNE~\citep{maaten2008visualizing}, utilizado para projetar representações internas em duas dimensões, facilitando a visualização da separação entre classes.
O modelo é relativamente leve no treinamento, com 1,34M de parâmetros ajustáveis, mas sua dependência de um backbone com 83M de parâmetros pode impactar negativamente o desempenho na fase de~inferência.

O \textit{\underline{CFM}}~\citep{luo:24} adota \textit{triplet learning} e mineração de amostras críticas para fomentar representações discriminativas e agnósticas ao domínio.
Alcançou alta transferibilidade, com AUC média de 0,84, e demonstrou robustez notável, apresentando uma queda média de apenas 3,71\% em sete tipos de perturbações visuais. Sua explicabilidade, assim como nos métodos anteriores, é limitada ao uso de Grad-CAM. Em termos de eficiência, a utilização de dois \textit{encoders} e amostragem por \textit{triplets} tende a elevar o custo computacional; no entanto, o uso do backbone EfficientNet-B4, com aproximadamente 19M de parâmetros, contribui para manter uma eficiência~moderada.

O \textit{\underline{TruthLens}}~\citep{kundu:25} se destaca por adotar uma abordagem multimodal, combinando o modelo visual DINOv2 com o LLM PaliGemma2 para gerar explicações em linguagem natural.
Alcançou acurácia média de 0,94 em três conjuntos de dados não vistos, evidenciando alto desempenho em termos de transferibilidade.
No entanto, sua avaliação de robustez foi limitada a cenários com compressão.
É o único método analisado com explicabilidade integrada, validada por métricas de linguagem como BLEU e ROUGE. 
Quanto à eficiência, o uso combinado de dois modelos fundacionais de grande porte impõe elevada complexidade computacional, o que pode limitar sua aplicação em tempo real ou em sistemas de larga~escala.

A Tabela~\ref{tab:robustez} apresenta os valores de robustez $R$ calculados para cada método, conforme a Equação~\ref{eq:robustez}. A métrica considera as médias de AUC ou acurácia dos detectores sob três tipos de perturbações: compressão~($Score_{\text{comp}}$), ruído~($Score_{\text{perturb}}$) e ataques adversariais~($Score_{\text{adv}}$).
Nos casos em que os testes não foram realizados, o valor 0 foi~adotado.

\begin{table}[!htb]
\centering
\caption{Comparação dos métodos quanto aos fatores de robustez.}
\setlength{\tabcolsep}{7pt}
\vspace{-2mm}

\label{tab:robustez}
\resizebox{0.675\linewidth}{!}{
\begin{tabular}{lccccc}
\toprule
Método & Métrica & $Score_{\text{comp}}$ & $Score_{\text{perturb}}$ & $Score_{\text{adv}}$ & $R$ \\
\midrule
SCLoRA     & AUC & 0{,}70 & 0{,}0   & 0{,}0 & 0{,}23 \\
OSDFD      & AUC & 0{,}79 & 0{,}87   & 0{,}0 & 0{,}55 \\
CFM       & AUC & 0{,}93 & 0{,}80   & 0{,}0 & 0{,}58 \\
FrePGAN & Acurácia & \textbf{0{,}99} & \textbf{0{,}97}   & 0{,}0 & \textbf{0{,}65} \\
TruthLens & Acurácia & 0{,}94 & 0{,}0   & 0{,}0 & 0{,}31 \\
\bottomrule
\end{tabular}
}
\end{table}

A Tabela~\ref{tab:resultados} consolida os escores de confiabilidade global ($SCG$), calculados conforme o Algoritmo~\ref{alg:confiabilidade}, bem como os valores individuais obtidos por cada método nos quatro pilares que compõem o framework: transferibilidade ($T$), robustez ($R$), interpretabilidade ($I$) e eficiência computacional ($E$).

\begin{table}[!htb]
\centering
\caption{Comparação dos métodos em relação aos pilares de confiabilidade.}
\label{tab:resultados}
\vspace{-2mm}
\resizebox{0.625\linewidth}{!}{
\begin{tabular}{lcccccc}
\toprule
Método & Métrica & $T$ & $R$ & $I$ & $E$ & $SCG$\\
\midrule
OSDFD      & AUC &  0{,}82 & 0{,}55 & 0{,}50 & 0{,}62 & 0{,}62\\
SCLoRA     & AUC & 0{,}72 & 0{,}23 & 0{,}20 & 0{,}60 & 0{,}44\\
CFM        & AUC & 0{,}84 & 0{,}58 & 0{,}50 & \textbf{0{,}80} & \textbf{0{,}68}\\
FrePGAN    & Acurácia & 0{,}76 & \textbf{0{,}65} & 0{,}30 & 0{,}58 & 0{,}57\\
TruthLens  & Acurácia & \textbf{0{,}94} & 0{,}31 & \textbf{1{,}00} & 0{,}00 & 0{,}56\\
\bottomrule
\end{tabular}
}
\end{table}

\subsection{Discussão}

A análise comparativa dos cinco métodos avaliados revela avanços significativos na detecção de deepfakes, mas também expõe fragilidades estruturais recorrentes, especialmente à luz dos quatro pilares de confiabilidade.
Apesar da \textit{transferibilidade} ter se tornado um foco crescente em abordagens recentes, com modelos como TruthLens e CFM alcançando acurácia média superior a 88\% em conjuntos de dados não vistos, observa-se uma persistente dependência de características específicas dos dados de treinamento, o que ainda limita a generalização plena desses métodos em cenários abertos.

No que diz respeito à \textit{robustez}, a maioria dos métodos avaliou seu desempenho frente a perturbações como compressão e ruído; no entanto, nenhum dos cinco métodos contemplou testes contra ataques adversariais --~uma lacuna crítica apontada por \cite{abdullah:24}, sobretudo em contextos de segurança.
Mesmo modelos com bom desempenho sob perturbações convencionais, como o FrePGAN e o CFM, ainda demonstram vulnerabilidade potencial diante desse tipo de~ameaça.

O pilar da \textit{interpretabilidade} apresenta disparidades ainda mais acentuadas.
Dentre os métodos avaliados, apenas o TruthLens incorpora um mecanismo explicativo integrado ao processo decisório, fornecendo descrições textuais que justificam a detecção.
Os demais métodos, quando oferecem alguma forma de explicação, limitam-se a visualizações básicas com técnicas como Grad-CAM ou t-SNE, sem uma análise interpretativa mais profunda.
Essa ausência de mecanismos explicativos pode comprometer a adoção de detectores em contextos forenses e jurídicos, nos quais as decisões precisam ser justificáveis e passíveis de auditoria~\citep{wang:24:deepfake}.

Com relação à \textit{eficiência computacional}, observa-se um trade-off recorrente: modelos altamente explicáveis (como o TruthLens) ou com forte capacidade de generalização (como o OSDFD) apresentam elevado custo de inferência, em razão do uso de arquiteturas fundacionais ou transformers de grande porte. Entre os métodos avaliados, apenas o CFM demonstra um equilíbrio razoável entre desempenho, robustez e eficiência, adotando um backbone intermediário com aproximadamente 19M de~ parâmetros.

Embora este trabalho forneça um panorama abrangente, reconhece-se que a comparação entre os métodos pode ser influenciada pela diversidade dos datasets utilizados nos experimentos originais e pela variação no número e tipo de testes de robustez realizados.
\blue{Essa limitação evidencia a necessidade de métricas padronizadas para avaliação da confiabilidade, que vão além de indicadores tradicionais como acurácia ou AUC, mas que incorporem uma visão mais ampla e multidimensional da confiabilidade dos~modelos.}

\section{Conclusões}
\label{sec:conclusoes}

Este trabalho apresentou uma análise crítica da confiabilidade de detectores de deepfakes, propondo um framework de avaliação baseado em quatro pilares fundamentais: transferibilidade, robustez, interpretabilidade e eficiência computacional.
A aplicação desse framework a cinco métodos do estado da arte permitiu identificar avanços relevantes, especialmente na capacidade de generalização para dados não vistos e na mitigação de artefatos de geração específicos.
Ainda assim, persistem limitações estruturais que comprometem a adoção confiável desses detectores em cenários~reais.

Entre os principais desafios identificados está a carência de testes contra ataques adversariais, o que fragiliza a robustez dos modelos diante de ameaças propositalmente elaboradas para ``enganá-los''.
A interpretabilidade também permanece como um gargalo considerável, com apenas um dos métodos avaliados oferecendo explicações integradas em linguagem natural.
Quanto à eficiência computacional, observou-se um trade-off entre complexidade e desempenho, que pode limitar a aplicabilidade dos modelos em tempo real ou em dispositivos com restrições de hardware.

Como próximos passos, propõe-se a ampliação do número de modelos analisados, com a reprodução dos experimentos em um ambiente padronizado, utilizando os mesmos conjuntos de dados, protocolos de teste e métricas de avaliação.
Acreditamos que essa padronização permitirá comparações mais justas entre os métodos e favorecerá o desenvolvimento de detectores realmente confiáveis, transparentes e acessíveis.

\section*{Agradecimentos}

Este trabalho foi financiado com recursos do Programa de Excelência Acadêmica~(PROEX) da Coordenação de Aperfeiçoamento de Pessoal de Nível Superior~(CAPES) – Brasil.
Agradecemos o apoio institucional, essencial tanto para a participação no evento quanto para o desenvolvimento desta pesquisa.

\bibliographystyle{sbc}
\bibliography{bibtex-short}

\end{document}